\documentclass{article}
\usepackage[T1]{fontenc}
\usepackage{spconf,amsmath,amssymb,graphicx,hyperref,bm,booktabs}
\usepackage{tikz}
\usetikzlibrary{arrows.meta,fit,backgrounds,positioning}
\title{PULSEFLOW: PPG COUNTERFACTUAL GENERATION VIA LATENT TRANSPORT}

\name{Hung Manh Pham$^{1}$, Dong Ma$^{2}$, Bin Zhu$^{1}$, Pan Zhou$^{1}$}

\address{
$^{1}$Singapore Management University, Singapore\\
$^{2}$University of Cambridge, United Kingdom\\
\texttt{hm.pham.2023@phdcs.smu.edu.sg, dm878@cam.ac.uk}\\
\texttt{binzhu@smu.edu.sg, panzhou@smu.edu.sg}
}

\begin{document}
\maketitle

\begin{abstract}
Photoplethysmography (PPG) has become an important modality for continuous cardiovascular monitoring, including atrial fibrillation (AF) detection. However, labeled AF recordings remain limited in many clinical settings, making model adaptation difficult when only limited target data are available. Generative modeling offers a natural way to alleviate this scarcity by synthesizing additional AF signals. Existing approaches, however, mainly generate samples that match the target condition without explicitly modeling how an observed source recording should be transformed, making it difficult to leverage abundant source recordings from a specific population or cohort for targeted augmentation. We introduce \textbf{PulseFlow}, a source-conditioned counterfactual generation framework that combines conditional representation learning with invertible latent transport to edit cardiac rhythm while retaining information from the source. Experiments across two clinical cohorts demonstrate effective rhythm transformation, measurable source correspondence, and improved AF classification under limited labels.

\end{abstract}


\section{Introduction}
\label{sec:intro}
Photoplethysmography (PPG) provides a low-cost and non-invasive means of continuously monitoring cardiovascular activity and has been widely used for downstream tasks such as heart rate estimation~\cite{meier2024wildppg}, stress assessment~\cite{wesad}, blood pressure estimation~\cite{gonzalez2023benchmark}, and atrial fibrillation (AF) detection~\cite{pereira2020review}. More recently, large-scale PPG foundation models (FMs)~\cite{papagei,anyppg,cardiostate} have shown strong transferability across these tasks. Despite this progress, adapting such models to a new clinical cohort or target population often still requires locally labeled data. In many clinical settings, acquiring reliable labels is costly, while important conditions such as AF may be under-represented. Generative modeling therefore provides a promising way to augment scarce physiological data.

Existing PPG generation methods~\cite{ding2023log,10892013,Jaiantilal2024Advantages} have shown value for downstream applications, but largely focus on generating signals that match the population-level characteristics of a target condition. They do not explicitly model how a particular observed recording should be transformed under that condition. This distinction is especially relevant for PPG, whose waveform characteristics exhibit substantial inter-individual variability and are associated with demographic factors such as age and sex~\cite{moscato2022quality}. When AF labels are scarce, abundant sinus-rhythm (SR) recordings from the target population therefore provide useful source information that conventional target-condition generation may not fully exploit. This motivates a source-conditioned counterfactual question: \textit{what would an observed SR recording look like if the cardiac rhythm were changed to AF while information associated with the source were retained?}

Structural causal models (SCMs) provide a principled formulation for counterfactual generation through abduction, intervention, and prediction~\cite {pawlowski_cf}: source information is first inferred from the observed recording, cardiac rhythm is then intervened on, and the resulting state is mapped back to the waveform space. A key challenge, however, is that matching the SR and AF distributions alone does not uniquely determine how a particular source state should correspond to a target state. To address this ambiguity, we propose \textbf{PulseFlow}, which combines conditional PPG representation learning with invertible latent transport. PulseFlow maps source and target states through a shared latent space while retaining source-derived information, with weakly paired SR-AF recordings further guiding cross-rhythm correspondence. At inference, the learned transport enables abduction, rhythm intervention, and counterfactual prediction from an observed source recording.

\noindent\textbf{Contributions.}
\textbf{(i)} We formulate PPG rhythm transformation as a source-conditioned counterfactual generation problem, aiming to modify cardiac rhythm while retaining information from the observed source;
\textbf{(ii)} We develop an invertible latent transport framework that preserves source-derived information and leverages weakly paired SR-AF recordings to guide cross-rhythm correspondence;
\textbf{(iii)} experiments across two clinical cohorts demonstrate effective rhythm transformation, measurable source correspondence, and downstream benefits for AF classification under limited labeled data.

\begin{figure*}[tbp]
\centering
\includegraphics[width=\linewidth]{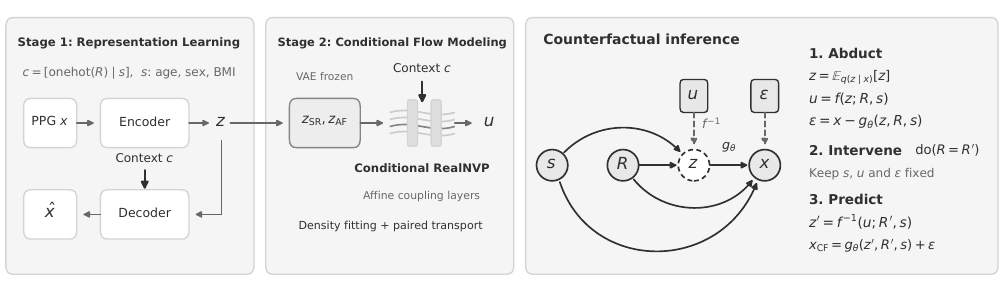}
\caption{Overview of PulseFlow: learning representation, conditional flow modeling, and counterfactual inference.}
\label{fig:overview}
\label{fig:dag}
\end{figure*}

\section{Method}
\label{sec:method}

\subsection{Overview}

PulseFlow consists of two stages, as illustrated in Fig.~\ref{fig:overview}. Stage~1 learns a compact representation of PPG waveforms together with a conditional decoder, while Stage~2 keeps this representation fixed and learns an invertible latent transport conditioned on cardiac rhythm and demographic context. During inference, an observed recording is mapped to its latent state, the rhythm condition is changed through the learned transport, and the resulting state is decoded into a source-conditioned counterfactual waveform.

\subsection{Structural Setting}
\label{ssec:scm}

We denote a 30\,s PPG segment by $\bm{x}\in\mathbb{R}^{3750}$, with context $\bm{c}=[\mathrm{onehot}(R)\,|\,\bm{s}]$, where $R\in\{\mathrm{SR},\mathrm{AF}\}$ indicates cardiac rhythm and $\bm{s}$ contains age, sex, and body mass index (BMI). The bottom panel of Fig.~\ref{fig:overview} shows the assumed structural model. Its observed variables are $R$, $\bm{s}$, and $\bm{x}$, while $\bm{z}\in\mathbb{R}^{128}$ denotes a latent waveform state. We treat $R$ and $\bm{s}$ as given and model neither their causes nor the relation between them. The edges $R\rightarrow\bm{z}$ and $\bm{s}\rightarrow\bm{z}$ describe how rhythm and demographics relate to the latent state, while $\bm{z}\rightarrow\bm{x}$ describes how this state is expressed in the waveform. The direct edges $R\rightarrow\bm{x}$ and $\bm{s}\rightarrow\bm{x}$ allow the same latent state to be expressed differently across contexts. We parameterize these mechanisms as
\begin{align}
\bm{z}&=f^{-1}(\bm{u};R,\bm{s}),\qquad \bm{u}\sim\mathcal{N}(\bm{0},\bm{I}),\label{eq:sz}\\
\bm{x}&=g_\theta(\bm{z},R,\bm{s})+\bm{\varepsilon},\label{eq:sx}
\end{align}
where $f$ is a conditional invertible flow and $g_\theta$ is a conditional waveform decoder, both receiving $(R,\bm{s})$ through $\bm{c}$. Under the observed rhythm $R$, the corresponding latent state and waveform are denoted by $\bm{z}$ and $\bm{x}$; replacing $R$ with a target rhythm $R'$ yields the counterfactual state $\bm{z}'$ and waveform $\bm{x}_{\mathrm{CF}}$. Following deep structural causal models~\cite{pawlowski_cf}, inference retains the source demographics and source-derived latent variables while intervening on rhythm. The resulting waveform is therefore a model-implied counterfactual under the assumed structural model rather than a uniquely identified biological outcome.

\subsection{Stage 1: Representation Learning}
\label{ssec:biocvae}

Stage~1 learns the waveform mechanism $g_\theta$ in Eq.~(\ref{eq:sx}) together with an encoder that infers $\bm{z}$ using a conditional variational autoencoder. A residual CNN encoder receives only $\bm{x}$ and parameterizes $q(\bm{z}\mid\bm{x})$, while a FiLM-conditioned decoder~\cite{film_perez} reconstructs $\bm{x}$ from $\bm{z}$ and $\bm{c}$. Training minimizes
\begin{equation}
\mathcal{L}_{\mathrm{VAE}}
=
\mathcal{L}_{\mathrm{rec}}
+
\beta\mathcal{L}_{\mathrm{KL}}
+
\lambda_{\mathrm{vital}}\mathcal{L}_{\mathrm{vital}},
\label{eq:vae}
\end{equation}
where $\mathcal{L}_{\mathrm{rec}}$ is the waveform reconstruction MSE, $\mathcal{L}_{\mathrm{KL}}$ regularizes the posterior toward a standard Gaussian, and $\mathcal{L}_{\mathrm{vital}}$ trains an auxiliary head to predict normalized mean RR from $\bm{z}$, with $\beta=0.08$ after warm-up and $\lambda_{\mathrm{vital}}=0.2$. The RR objective encourages the latent representation to retain beat-timing information relevant to distinguishing SR from AF. The encoder and decoder are then frozen, and the posterior mean of $q(\bm{z}\mid\bm{x})$ serves as the latent state for Stage~2.

\subsection{Stage 2: Conditional Flow Modeling}
\label{ssec:rhythmpgm}

Stage~2 learns the latent mechanism $f$ in Eq.~(\ref{eq:sz}) using an adapted conditional RealNVP~\cite{dinh2017} with eight affine coupling layers. The flow maps latent states to a shared base space while conditioning on rhythm and demographic context. By invertibility, its conditional density is
\begin{equation}
\log p(\bm{z}\mid\bm{c})
=
\log p_{\bm{u}}\!\left(f(\bm{z};\bm{c})\right)
+
\log\left|\det\frac{\partial f}{\partial\bm{z}}\right|.
\label{eq:density}
\end{equation}
We first fit the context-conditioned latent distributions by minimizing the negative log-likelihood,
\begin{equation}
\mathcal{L}_{\mathrm{NLL}}
=
-\frac{1}{N}\sum_{i=1}^{N}\log p(\bm{z}_i\mid\bm{c}_i).
\end{equation}
Likelihood fitting alone, however, does not determine which AF state should correspond to a particular SR state, since multiple transports can reproduce the same target distribution. We therefore define
\[
T_{a\rightarrow b}(\bm{z})
=
f^{-1}\!\left(f(\bm{z};\bm{c}_a);\bm{c}_b\right),
\]
which recovers $\bm{u}$ under the source context and reuses it under the target context. Same-patient recordings observed under both rhythms are then used to guide this cross-rhythm correspondence:
\begin{equation}
\mathcal{L}_{\mathrm{cf}}
=
\frac{1}{|\mathcal{D}|d}
\sum_{(a,b)\in\mathcal{D}}
\left\|
T_{a\rightarrow b}(\bm{z}_a)-\bm{z}_b
\right\|_2^2,
\label{eq:consist}
\end{equation}
where $d=128$ and $\mathcal{D}$ contains both SR-to-AF and AF-to-SR directions for each sampled same-patient pair. Stage~2 is trained by minimizing
$\mathcal{L}_{\mathrm{NLL}}+\mathcal{L}_{\mathrm{cf}}$.
Because the paired recordings are acquired at different times and may differ in beat phase and clinical state, they serve as weak supervision for learning cross-rhythm correspondence rather than as exact counterfactual targets.

\subsection{Counterfactual Inference}
\label{ssec:inference}

Given an observed recording $(\bm{x},\bm{c})$ without a paired target, generation follows abduction, intervention, and prediction~\cite{pawlowski_cf}. During \textbf{abduction}, the encoder provides the posterior mean $\bm{z}$, the flow recovers $\bm{u}=f(\bm{z};\bm{c})$, and the source reconstruction residual is computed as
$\bm{\varepsilon}=\bm{x}-g_\theta(\bm{z},\bm{c})$.
During \textbf{intervention}, only the rhythm component is changed, giving $\bm{c}'=[\mathrm{onehot}(R')\,|\,\bm{s}]$, while the source demographics, $\bm{u}$, and $\bm{\varepsilon}$ are retained. During \textbf{prediction}, the latent state is transported under the target context,
$\bm{z}'=f^{-1}(\bm{u};\bm{c}')$, and decoded as
\begin{equation}
\bm{x}_{\mathrm{CF}}
=
g_\theta(\bm{z}',\bm{c}')
+
\bm{\varepsilon}.
\label{eq:prediction}
\end{equation}
Retaining $\bm{u}$ ties the transported state to the observed source, while $\bm{\varepsilon}$ restores waveform information not captured through the VAE bottleneck. Flow inversion is exact under the learned model, whereas VAE encoding remains approximate.

\begin{table}[t]
\centering\small\setlength{\tabcolsep}{0.5pt}
\caption{Rhythm editing quality. Arrows mark proximity to real data ($\to$) or lower is better ($\downarrow$).}
\label{tab:main}
\begin{tabular}{lcccccc}
\toprule
& \textbf{Rhythm} & \multicolumn{4}{c}{\textbf{Realism}} & \textbf{RR match} \\
\cmidrule(lr){2-2} \cmidrule(lr){3-6} \cmidrule(lr){7-7}
\textbf{Method} & AF conv & FD & FD & SDNN  &  RMSSD & $W_1$ \\
& \%~$\to$ & 1~$\downarrow$ & 2~$\downarrow$ & ms~$\to$ & ms~$\to$ & ms~$\downarrow$ \\
\midrule
Real data & 92.6 & 0.49 & 0.14 & 120.5 & 166.7 & -- \\
\midrule
PPG-VAE~\cite{Jaiantilal2024Advantages} & 69.0 & 216.7 & 278.9 & 127.1 & 186.7 & 133.5 \\
Diff-SCM~\cite{sanchez2022diffcf} & 61.6 & 83.2 & 29.5 & 100.7 & 142.2 & 166.7 \\

Rectified flow~\cite{liu2022flow} & 95.3 & 58.2 & 20.8 & 142.1 & 189.7 & 126.3 \\
WGAN-GP~\cite{gulrajani2017wgangp} & 83.9 & 80.4 & 36.8 & 152.0 & 203.8 & 160.8 \\
Latent DDPM~\cite{wang2024ecg} & 94.8 & 56.4 & 21.3 & 139.6 & 190.2 & 119.8 \\
DenoisePPG-DM~\cite{10892013}  & 93.2 & 53.2 & 19.7 & 138.5 & 187.5 & \textbf{112.3} \\

\midrule
\textbf{PulseFlow} & \textbf{93.1} & \textbf{45.5} & \textbf{13.0} & \textbf{117.0} & \textbf{158.5} & 113.6 \\
\bottomrule
\end{tabular}

\end{table}

\section{Experimental Setup}
\label{sec:protocol}
\subsection{Datasets and baselines}
\label{ssec:data}

To facilitate large-scale learning of rhythm-dependent PPG variation, we leverage MIMIC-III-Ext-PPG~\cite{mimic3extppg}, which provides simultaneous PPG and ECG recordings from intensive care patients. We use 30\,s PPG segments at 125\,Hz and split 5,668 patients into training, validation, and test sets at a ratio of approximately 80/10/10\%, so that no patient appears in more than one set. The Stage~1 VAE is trained on approximately one million segments, and Stage~2 is fitted to the latent states of the same training segments. For $\mathcal{L}_{\mathrm{cf}}$, we additionally use 400 training patients recorded under both SR and AF. 

For baselines, we consider general generative models adapted for time series data, including Diff-SCM~\cite{sanchez2022diffcf}, a rectified flow~\cite{liu2022flow}, and a latent DDPM~\cite{wang2024ecg}, as well as related models used for PPG signal generation, namely PPG-VAE~\cite{Jaiantilal2024Advantages}, WGAN-GP~\cite{gulrajani2017wgangp}, and DenoisePPG-DM~\cite{10892013}. All baselines are adapted to PPG generation under the target AF condition using the same dataset and consistent evaluation setup. Depending on their implementation nature, the models are conditioned either on the target rhythm alone or additionally on the observed SR source, with source-conditioned methods using the same paired SR-AF training data where applicable. Diffusion and flow baselines are sampled with 100 steps, whereas PPG-VAE and DenoisePPG-DM generate waveforms directly.

\subsection{Evaluation protocol}
\label{ssec:metrics}

To evaluate the generated counterfactuals, we consider three complementary aspects: target-rhythm conversion, distributional similarity, and rhythm correspondence. First, conversion measures the fraction of generated signals classified as AF by a PPG foundation model~\cite{papagei}. We then assess distributional similarity using the Fr'echet distance to real AF signals in the AnyPPG~\cite{anyppg} and CardioState~\cite{cardiostate} representation spaces, as FD1 and FD2 correspondingly, together with standard beat-interval statistics, including SDNN and RMSSD, where values closer to real AF recordings indicate better physiological agreement~\cite{pereira2020review}. Finally, rhythm correspondence is measured by the patient-level Wasserstein-1 distance ($W_1$) between the beat-interval distributions of generated and observed AF recordings. 

\subsection{Augmentation with scarce labels}
\label{ssec:aug_setup}

To examine whether the generated counterfactuals are useful beyond the MIMIC cohort, we further evaluate data augmentation on VitalDB~\cite{vitaldb}, a surgical dataset unseen during generator training, using the provided SR and AF labels. Since AF patients are rare, we keep up to 30 windows per AF patient and 6 per SR patient. An AF classifier~\cite{papagei} is trained with 1, 2, and 4 AF patients (Settings~1 to 3) and a fixed set of 60 SR patients. SR recordings from 60 further patients serve as generation sources, and every baseline method adds the same number of synthetic AF windows. Finally, testing uses 8 AF and 60 SR patients, which results in about 40\% AF windows over five patient-level partitions.


\section{Results}
\label{sec:results}

\subsection{Rhythm conversion and source correspondence}
\label{ssec:res_realism}

\begin{figure}[tbp]
\centering
\includegraphics[width=\columnwidth]{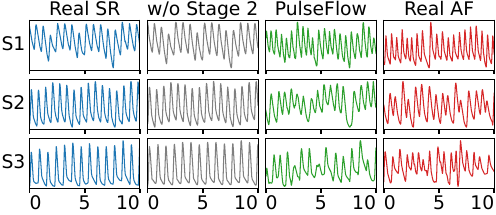}
\caption{SR-to-AF counterfactuals for three test patients.}
\label{fig:waves}
\end{figure}

Table~\ref{tab:main} evaluates whether the generated signals reach the target AF rhythm while remaining close to real AF recordings. PulseFlow achieves an AF conversion rate of 93.1\%, close to the 92.6\% observed for real AF. It also obtains the lowest FD among all generation methods in both AnyPPG and CardioState spaces, indicating better agreement with the real AF distribution. The generated rhythm statistics are similarly close to the real reference: PulseFlow achieves an SDNN of 117.0\,ms and an RMSSD of 158.5\,ms, compared with 120.5\,ms and 166.7\,ms for real AF. For patient-level RR correspondence, DenoisePPG-DM obtains a slightly lower $W_1$ than PulseFlow (112.3 vs.~113.6\,ms). Beyond these rhythm-level measures, we additionally assess source correspondence via patient retrieval in the CardioState space. PulseFlow achieves 17.2\% top-1 patient retrieval, close to the 19.7\% obtained from real cross-rhythm recordings and well above the 5\% chance level (20 subjects). Finally, Fig.~\ref{fig:waves} shows that decoding the source latent under the AF condition without Stage~2 largely retains the regular SR pattern, whereas PulseFlow produces less regular beat timing that more closely resembles the AF recordings.

\begin{table}[t]
\centering\small
\caption{VitalDB augmentation under scarce AF labels. AUROC (\%), mean $\pm$ SD over 5 partitions.}
\label{tab:incontext}
\resizebox{\columnwidth}{!}{\small
\begin{tabular}{lccc}
\toprule
\textbf{Augmentation} & \textbf{Setting 1} & \textbf{Setting 2} & \textbf{Setting 3} \\
\midrule
Real only & 67.6{\scriptsize$\pm$8.9} & 76.2{\scriptsize$\pm$6.5} & 81.4{\scriptsize$\pm$5.1} \\
\midrule
PPG-VAE~\cite{Jaiantilal2024Advantages} & 68.3{\scriptsize$\pm$9.6} & 77.4{\scriptsize$\pm$5.5} & 85.2{\scriptsize$\pm$5.7} \\

Diff-SCM~\cite{sanchez2022diffcf} & 72.6{\scriptsize$\pm$8.7} & 79.5{\scriptsize$\pm$5.8} & 85.5{\scriptsize$\pm$3.5} \\

Rectified flow~\cite{liu2022flow} & 76.2{\scriptsize$\pm$5.4} & 82.0{\scriptsize$\pm$2.6} & 87.0{\scriptsize$\pm$2.6} \\

WGAN-GP~\cite{gulrajani2017wgangp} & 73.7{\scriptsize$\pm$7.9} & 80.1{\scriptsize$\pm$3.6} & 85.8{\scriptsize$\pm$5.5} \\

Latent DDPM~\cite{wang2024ecg} & 77.2{\scriptsize$\pm$7.2} & 82.4{\scriptsize$\pm$3.5} & 86.9{\scriptsize$\pm$3.9} \\

DenoisePPG-DM~\cite{10892013} & 78.0{\scriptsize$\pm$6.4} & 83.8{\scriptsize$\pm$5.2} & 87.5{\scriptsize$\pm$3.3} \\

\midrule
\textbf{PulseFlow} & \textbf{83.6{\scriptsize$\pm$3.2}} & \textbf{86.7{\scriptsize$\pm$5.1}} & \textbf{87.9{\scriptsize$\pm$5.6}} \\
\bottomrule
\end{tabular}
}
\end{table}

\subsection{Augmentation performance}
\label{ssec:res_utility}

We next examine whether the generated signals provide useful augmentation on an unseen clinical cohort. As shown in Table~\ref{tab:incontext}, PulseFlow achieves the highest mean AUROC across all three scarce-label settings. Specifically, the largest improvement occurs in Setting~1, where only one AF patient is available: augmentation with PulseFlow increases AUROC from 67.6\% with real data alone to 83.6\%, compared with 78.0\% for DenoisePPG-DM, the strongest baseline in this setting. As additional AF labels become available, PulseFlow reaches 86.7\% and 87.9\% in Settings~2 and~3, while its margin over DenoisePPG-DM narrows from 5.6 to 2.9 and 0.4 percentage points. These results indicate that the augmentation benefit is most pronounced when real AF supervision is highly limited.

\begin{table}[t]
\centering\small
\setlength{\tabcolsep}{2pt}
\caption{Ablation experiments. Arrows indicate proximity to real data ($\to$), lower is better ($\downarrow$); Aug denotes Setting~1 augmentation.}
\label{tab:ablation}
\begin{tabular}{lcccccc}
\toprule
\textbf{Variant} & AF conv & FD & SDNN & RMSSD & $W_1$ & Aug \\
& \%~$\to$ & 1~$\downarrow$ & ms~$\to$ & ms~$\to$ & ms~$\downarrow$ & \%~$\uparrow$ \\
\midrule
\quad w/o Stage 2 & 5.5 & 178.5 & 28.2 & 35.0 & 162.6 & 68.4 \\
\quad w/o $\bm{\varepsilon}$ & 93.6 & 47.5 & 115.0 & 155.1 & \textbf{112.8} & 75.0 \\
\quad w/o $\mathcal{L}_{\mathrm{cf}},\bm{\varepsilon}$ & 87.9 & 52.8 & 128.8 & 176.6 & 142.3 & 72.0 \\
\quad w/o $\mathcal{L}_{\mathrm{vital}}$ & 92.1 & 67.4 & 126.5 & 175.7 & 124.4 & 81.0 \\
\midrule
\textbf{PulseFlow} & \textbf{93.1} & \textbf{45.5} & \textbf{117.0} & \textbf{158.5} & 113.6 & \textbf{83.6} \\
\bottomrule
\end{tabular}

\end{table}

\subsection{Component ablations}
\label{ssec:res_faithful}

Table~\ref{tab:ablation} examines the contribution of PulseFlow's main components. Without Stage~2, directly decoding the source latent under the AF condition yields only 5.5\% AF conversion and substantially higher FD, showing that changing the decoder condition alone is insufficient to produce the rhythm transition. We can also observe that removing the residual $\bm{\varepsilon}$ has little effect on conversion or RR correspondence, but reduces augmentation AUROC from 83.6\% to 75.0\%, suggesting that retaining source waveform information benefits downstream training. We also examine paired transport supervision by comparing the ``w/o $\bm{\varepsilon}$'' and ``w/o $\mathcal{L}_{\mathrm{cf}},\bm{\varepsilon}$'' variants. With residual removal fixed, removing $\mathcal{L}_{\mathrm{cf}}$ decreases AF conversion from 93.6\% to 87.9\% and increases $W_1$ from 112.8 to 142.3\,ms, supporting its role in guiding cross-rhythm correspondence. Finally, removing $\mathcal{L}_{\mathrm{vital}}$ increases FD from 45.5 to 67.4, showing the benefit of retaining rhythm-related information in the latent space.

\section{Conclusion}
\label{sec:conclusion}
We introduced PulseFlow for source-conditioned PPG counterfactual generation, using conditional representation learning with invertible latent transport. The results demonstrate that PulseFlow can produce effective rhythm transformations while maintaining source correspondence, and that the generated signals provide practical value for AF classification when labeled data are scarce. Future work will extend the framework to broader physiological factors and more explicit modeling of inter-individual PPG variation.

\bibliographystyle{IEEEbib}
\bibliography{refs}

\end{document}